\documentclass[11pt,letterpaper]{article}
\usepackage{emnlp2017}
\usepackage{times}
\usepackage{latexsym}
\usepackage{latexsym}
\usepackage{CJK}
\usepackage{graphicx}
\usepackage{multirow}
\usepackage{url}

\emnlpfinalcopy

\def\emnlppaperid{619}

\title{ Improving Social Media Text Summarization by Learning \\Sentence Weight Distribution}

\author{Jingjing Xu\\
MOE Key Laboratory of Computational Linguistics, Peking University\\
School of Electronics Engineering and Computer Science, Peking University\\
{jingjingxu}@pku.edu.cn}

\date{}

\begin{document}

\maketitle

\begin{abstract}

Recently, encoder-decoder models are widely used in social media text summarization. However, these models sometimes select noise words in irrelevant sentences as part of a summary by error, thus declining the performance. In order to inhibit irrelevant sentences and focus on key information, we propose an effective approach by learning sentence weight distribution. In our model, we build a multi-layer perceptron to predict sentence weights. During training, we use the ROUGE score as an alternative to the estimated sentence weight, and try to minimize the gap between estimated weights and predicted weights. In this way, we encourage our model to focus on the key sentences, which have high relevance with the summary. Experimental results show that our approach outperforms baselines on a large-scale social media corpus.

\end{abstract}

\section{Introduction}
\label{introduction}

 \begin{figure*}[!hbt]
  \centerline{\includegraphics[width=0.75\textwidth]{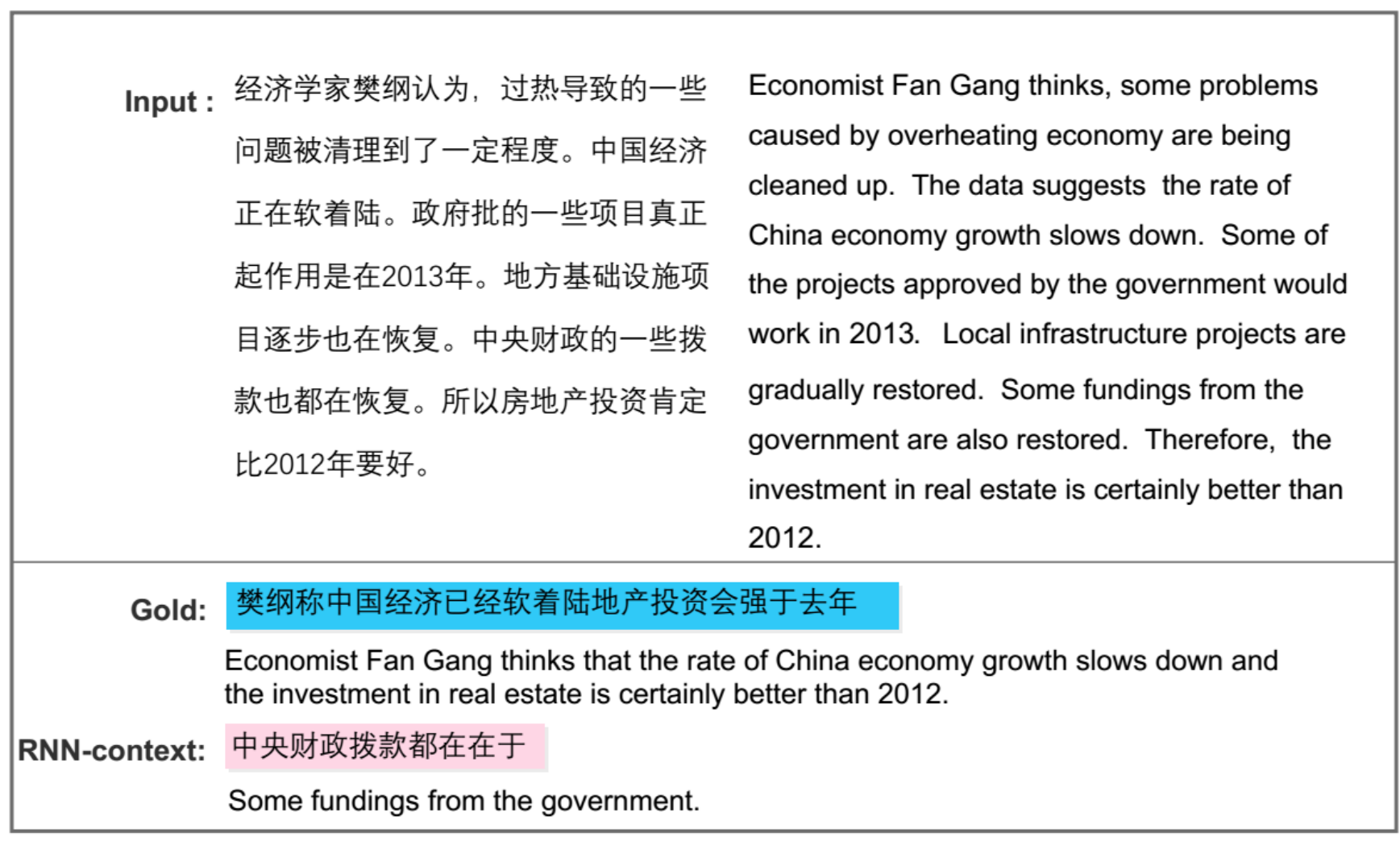}}
\caption{Illustration of the negative influence of noise words. ``RNN-context'' is the basic encoder-decoder model with the attention mechanism. The gold summary (shown in blue) is that the rate of China economy growth slows down and the investment in real estate is certainly better than 2012. However, the key words are unseen in the output of ``RNN-context'' while words (shown in pink) in irrelevant sentences are selected as part of a summary. }
\label{example}
\end{figure*}

Text summarization is an important task in natural language processing. It aims to understand the key idea of a document and generate a headline or a summary. Previous social media text summarization systems~\cite{RushCW15,HuCZ15} are mainly based on abstractive text summarization. Most of them belong to a family of encoder-decoders which have shown effective in many tasks, like machine translation~\cite{cho-EtAl,ChoMGBBSB14}. 

However, these models sometimes select noise words in irrelevant sentences as part of a summary by error. Figure \ref{example} gives an example of noise words generated by a state-of-the-art encoder-decoder model (``RNN-context'').   Unlike translation which requires encoding all information to ensure the accuracy, summarization tries to extract the most important information. Furthermore, only a small part of sentences convey the key information while the rest of sentences usually are useless. Thus, these unrelated sentences make it hard for encoder-decoder models to extract key information.

In order to address this issue, we propose a novel method by learning sentence weight distribution to encourage models focus on key sentences and ignore unimportant sentences. In our approach, we first design a multi-layer perceptron to predict sentence weights. Then, considering that ROUGE is a popular evaluation criterion for summarization, we estimate the gold sentence weights of training data by ROUGE scores between sentences and summaries.  During training, we design an end-to-end optimization method which minimizes the gap between predicted sentence weights and estimated sentence weights.





 We conduct experiments on a large-scale social media dataset. Experimental results show that our method outperforms competitive baselines. Besides, we do not limit our method to any specific neural network, it can be extended to any sequence-to-sequence model.

%
%








\section{Related Work}

Summarization approaches can be divided into two typical categories: extractive summarization
~\cite{radev2004mead,aliguliyev2009new,woodsend2010automatic,ferreira2013assessing,cheng-lapata}
 and abstractive summarization~\cite{knight2002summarization,bing-EtAl,RushCW15,HuCZ15,GuLLL16}. For extractive summarizations, most works usually select several sentences from a document as a summary or a headline. For abstractive summarization, most works usually encode a document into an abstractive representation and then generate words in a summary one by one.  Most social media summarization systems belong to abstractive text summarizaition. Generally speaking, extractive summarization achieves better performance than abstractive summarization for long and normal documents. However, extractive summarization is not suitable for social media text which are full of noises and very short. 

Neural abstractive text summarization is a newly proposed method and has become a hot research topic in recent years. Unlike the traditional summarization systems which consist of many small sub-components that are tuned separately~\cite{knight2002summarization,erkan2004lexrank,moawad2012semantic}, neural abstractive text summarization attempts to build and train a single, large neural network that reads a document and outputs a correct summary. Rush et al.~\shortcite{RushCW15} first introduced the encoder-decoder framework with the attention mechanism to abstractive text summarization. Bing et al.~\shortcite{bing-EtAl} proposed an abstraction-based multi-document summarization framework which can construct new sentences by exploring
more fine-grained syntactic units than sentences. Gu et al.~\shortcite{GuLLL16} proposed a copy mechanism to address the problem of unknown words. Nallapati et al.~\shortcite{NallapatiZSGX16} proposed several novel models to address critical problems in summarization.


\section{Proposed Model}

Our method is based on the basic encoder-decoder framework proposed by Cho et al.~\shortcite{ChoMGBBSB14} and Sutskever et al.~\shortcite{sutskever2014sequence}. 

Section 3.1 introduces how to estimate sentence weight distribution in detail. Section 3.2 describes how to generate the representation of sentence weight distribution. Section 3.3 shows how to incorporate estimated sentence weights and predicted sentence weights in training.

\subsection{Estimating Sentence Weight Distribution}



Assume we are provided with a summary $y$ and a document $D=d_{1},...,d_{n}$ where $n$ is the number of sentences. The first step of our method is to compute the distribution of sentence weights for training data as
\begin{eqnarray}
w=\{w_{1},...,w_{i},...,w_{n}\}
\end{eqnarray}
where $w_{i}$ is computed as
\begin{eqnarray}
w_{i}=\frac{exp(e_{i})}{\sum_{j=1}^{n}exp(e_{j})}
\end{eqnarray}
and $e_{i}$ is computed as
\begin{eqnarray}
e_{i}=ROUGE(d_{i},y)
\end{eqnarray}
where ROUGE is the evaluation metric to judge the quality of predicted summaries. 


\subsection{Representation of Sentence Weight Distribution}

In our model, we first produce sentence weight distribution $w^{'}$ over all sentences. The computation is
based on the sentence embeddings $s$ and the position embeddings of sentences $index(s)$ as

\begin{eqnarray}
w^{'}_{j}=\frac{exp(MLP([s_{j},index(s_{j})]) )}{\sum_{j^{'}=1}^{n}exp(MLP([s_{j^{'}},index(s_{j^{'}})])) }
\end{eqnarray}
where $[s_{j},index(s_{j})]$ denotes vector concatenation of $s_{j}$ and $index(s_{j})$; MLP refers to a
multi-layer perceptron. 
 $s_{j}$ is produced as
\begin{eqnarray}
s_{j}=\sum _{i\in sen2word(j)}(x_{i})
\end{eqnarray}
where $sen2word(j)$ returns all indexes of words which belong to the $j_{th}$ sentence.   
Then, the new output of an encoder part is
\begin{eqnarray}
\textbf{h}=\{h^{'}_{1},...,h^{'}_{i},...,h^{'}_{n}\}
\end{eqnarray}
where $n$ is the number of hidden states and $h^{'}_{i}$ is computed as
\begin{eqnarray}
h^{'}_{i}=w^{'}_{word2sen(i)}h_{i}
\end{eqnarray}
where $p_{word2sen(i)}$ returns the weight of sentence which $x_{i}$ belongs to, and $h_{i}$ is the output of RNN or Bi-LSTM used in an encoder. Then, the new output of an encoder $\textbf{h}$ is delivered to a decoder which produces a summary. 

 \begin{figure*}[!hbt]
  \centerline{\includegraphics[width=0.75\textwidth]{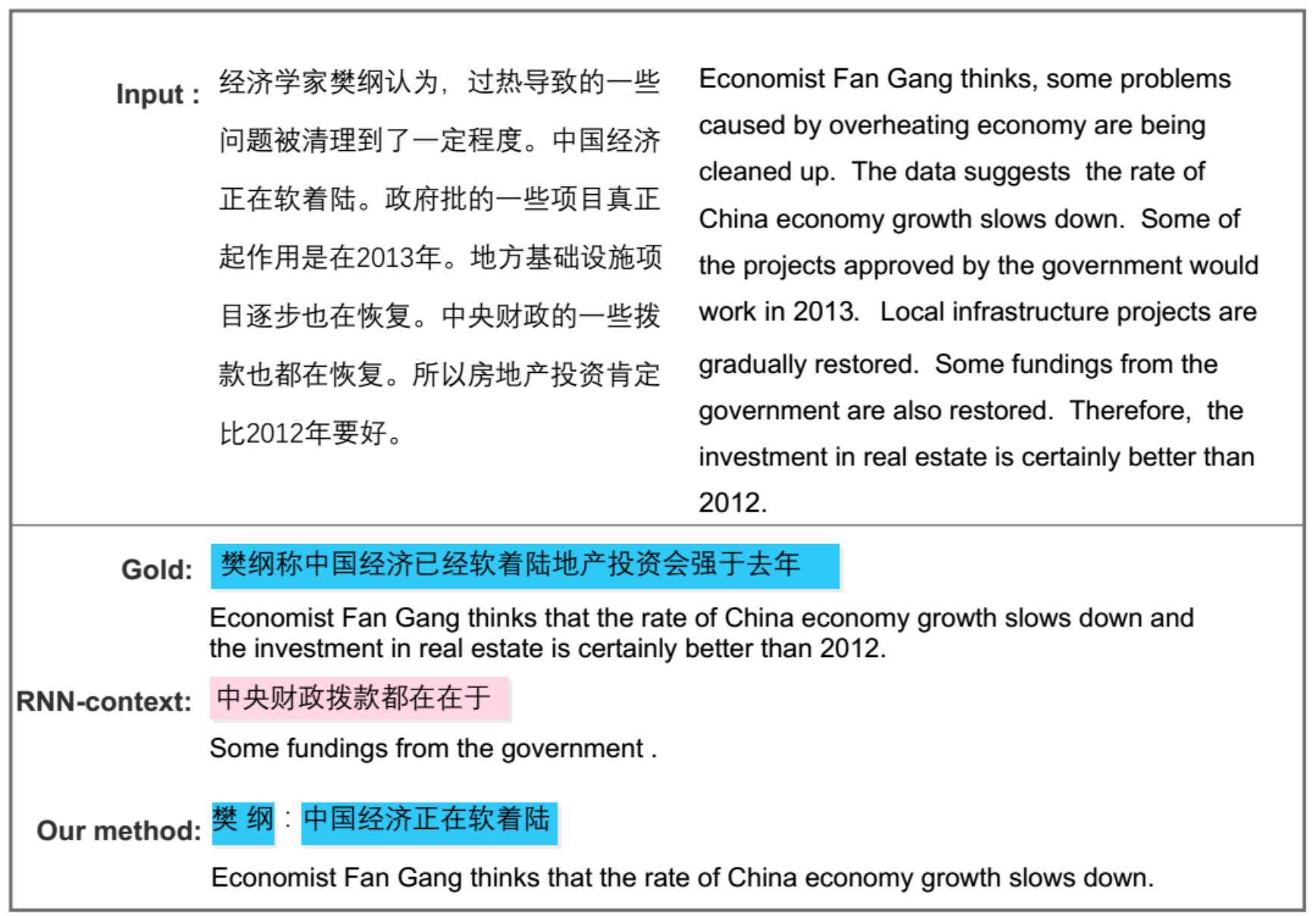}}
\caption{Comparisons of the predicted summaries between RNN-context and our method. The predicted summary (shown in blue) of RNN-context comes from an unimportant sentence. In contrast, the predicted summary of our method covers some key words (shown in purple). }
\label{examplebnn}
\end{figure*}

\subsection{Training}

Given the model parameter $\theta$ and an input text $x$, a corresponding summary $y$ and  sentence weight distribution $s$ (described in Section 3.1), the loss function is
\begin{eqnarray*}
J(\theta)=-\frac{1}{N}\sum_{i=1}^{N}(\sum_{j=1}^{m}{log(p(y_{i,j}|X_{i},\theta))}+\\ 
\lambda \sum_{j=1}^{k}{ w^{'}_{i,j}log(p(w_{i,j}|X_{i},\theta))})
\end{eqnarray*}
where $N$ is the batch size, $p(y_{i,j}|X_{i},\theta)$ is the conditional probability of the output word $y_{i,j}$ given source texts $X_{i}$, $ w^{'}_{i,j}$ is the predicted sentence weight and $p(w_{i,j}|X_{i},\theta)$ is the conditional probability of the sentence weight $w_{i,j}$ (descirbed in Section 3.1) given source texts $X_{i}$. 

%
%
%
%
\begin{table}
\centering
\begin{tabular}{|c|c|c|}
\hline
Train&Devlopment&Test\\
\hline
2,400,591&10,666&1,106\\
\hline
\end{tabular}
\caption{Details of LCSTS dataset. The size is given in number of pairs (short text, summary). }
\label{transfer1}
\end{table}

\begin{table}
\centering
\begin{tabular}{|c|c|c|c|}
\hline
\multicolumn{1}{|c|}{\multirow{1}{*}{Models}}&R-1&R-2&R-L\\



\hline
RNN~\cite{HuCZ15}&21.5&8.9&18.6\\
\textbf{+SWD}&\textbf{24.1}&\textbf{10.3}&\textbf{21.1}\\
\hline
\multirow{1}{*}{RNN-context}&\multirow{2}{*}{29.9}&\multirow{2}{*}{17.4}&\multirow{2}{*}{27.2}\\

~\cite{HuCZ15}&&&\\
\textbf{+SWD}&\textbf{32.0}&\textbf{19.0}&\textbf{29.4}\\
\hline

\hline

\end{tabular}

\caption{ROUGE scores (R-1:ROUGE-1; R-2: ROUGE-2; R-L: ROUGE-L) of the trained models computed on the test and development sets. ``RNN'' and ``RNN-context'' are two baselines. We refer our method as SWD.  }
\label{comneura}
\end{table}

%
%
%
%
%
%


\section{Experiments}
In this section, we evaluate our proposed approach on a social media dataset and report the performance of the models. Furthermore, we use a case to illustrate the improvement achieved by our approach.



\subsection{Dataset}

We use the large-scale Chinese short summarization dataset (LCSTS), which is provided by Hu et al.~\shortcite{HuCZ15}. This dataset is constructed from a famous Chinese social media called Sina Weibo\footnote{The place where a lot of popular Chinese media and organizations post news and information.}.  Based on the statistic data on the training set, we set the maximum number of sentences as 20 and the maximum length of a sentence as 150 in this paper. 


\subsection{Experimental Settings}

Following previous works and experimental results on the development set, we set hyper-parameters as follows. The character embedding dimension is 400 and the size of hidden state is 512. The parameter $\lambda$ is 0.01. All word embeddings are initialized randomly. We use the 1-layer encoder and the 1-layer decoder in this paper.

We use the minibatch stochastic gradient descent (SGD) algorithm to train our model. Each gradient is computed using a minibatch of 32 pairs (document, summary). Best validation accuracy is reached after 12k batches, which requires around 2 days of training. For evaluation, we use the ROUGE metric proposed by \cite{rough}. Unlike BLEU which includes various n-gram matches, there are several versions of ROUGE for different match lengths: ROUGE-1, ROUGE-2 and ROUGE-L.
Experiments are performed on a commodity 64-bit Dell Precision T7910 workstation with one 3.0 GHz 16-core CPU, 	 RAM and one Titan X GPU.

\subsection{Models}
We do not limit our method on specific neural network, it can be extended to any sequence-to-sequence model. In this paper, we evaluate our method on two types of baselines.
%
%

\textbf{RNN} We denote RNN as the basic sequence-to-sequence model, with a bi-LSTM encoder and a bi-LSTM decoder. It is a widely used framework. 
%
%

\textbf{RNN-context} RNN-context is a  sequence-to-sequence framework with the attention mechanism. 

\subsection{Results and Discussions}

We compare our approach with baselines, including RNN and RNN-context. The main results are shown in Table 1. It can be seen that our approach achieves ROUGE improvement over both baselines. In particular, SWD outperforms RNN-context by almost 2\% ROUGE-1 points. 




Finally we give an example summary as shown in Figure \ref{examplebnn}. This example is illustrated in Section \ref{introduction}, aimed to show the negative influence of unimportant words on extracting key information on RNN-context. RNN-context chooses some unimportant words as summary, like ``some fundings from the central government'' (shown in blue). In contrast, the outputs of our method contain some key words (shown in pink), like ``Fan Gang'', `` the rate of China economy growth slows down''. This example shows the effectiveness of our model on handling noise documents which are full of a number of irrelevant words. 


\section{Conclusion}

In this paper, we propose a novel method by learning sentence weight distribution to improve the performance of abstractive  summarization. The target is to make models focus on important sentences and ignore irrelevant sentences. The results on a large-scale Chinese social media dataset show that our approach outperforms competitive baselines. We also give the example which shows that the summary produced by our method is more relevant to the gold summary. Besides, our method can be extended to any sequence-to-sequence model. Word based seq2seq systems are potentially helpful to this task, cause the words can incorporate more meaningful information. In the future, we will try several word segmentation methods~\cite{SunZMTT09,SunWL12,Xu2016Dependency,Jingjing2017} to improve the system.

\nocite{MA2017,shumingma,SunZMTT13,SunLWL14,iclr2018}

\bibliography{emnlp2017}

\bibliographystyle{emnlp_natbib}

\end{document}